\pdfoutput=1
\PassOptionsToPackage{table,x11names}{xcolor}
\documentclass[11pt]{article}

\usepackage[final]{acl}
\usepackage{float}

\usepackage{amsmath}
\usepackage{textcomp}
\usepackage{multirow}
\usepackage{enumitem}
\usepackage{paralist}
\usepackage[most]{tcolorbox}
\usepackage{booktabs}

\usepackage{booktabs}
\usepackage{makecell}

\usepackage{nicematrix}

\usepackage{times}
\usepackage{latexsym}

\usepackage[T1]{fontenc}
\usepackage[utf8]{inputenc}

\usepackage{microtype}

\usepackage{graphicx}
\usepackage{colortbl}
\usepackage{array}
\usepackage{xcolor}
\definecolor{Rust}{RGB}{183,65,14}

\usepackage{inconsolata}

\usepackage{helvet}
\usepackage{graphicx}

\definecolor{mypurple}{RGB}{153,0,255}
\definecolor{myorange}{RGB}{240,123,0}
\definecolor{mygrey}{RGB}{100,100,100}

\usepackage{times}
\usepackage{latexsym}
\usepackage{enumitem}
\setdescription{itemsep=0pt,parsep=0pt,leftmargin=0.5cm}
\usepackage{pifont}
\usepackage{colortbl}
\usepackage[framemethod=tikz]{mdframed}

\usepackage{hyperref}
\hypersetup{
    colorlinks = true,
    allcolors = mypurple!40!black}

\newmdenv[roundcorner=4pt,linecolor=Rust,backgroundcolor=Rust!10,innerleftmargin=3pt,
innerrightmargin=3pt,innertopmargin=3pt,innerbottommargin=3pt]{mybox}

\title{TopicImpact: Improving Customer Feedback Analysis with\\ Opinion Units for Topic Modeling and Star-Rating Prediction}

\author{Emil Häglund \\
Department of Computing Science \\
  Umeå University, Sweden \\
  {\tt emilh@cs.umu.se} \\\And
  Johanna Björklund \\
Department of Computing Science \\
  Umeå University, Sweden \\
  {\tt johanna@cs.umu.se}}

\begin{document}
\maketitle
\begin{abstract}
 %We improve the extraction of insights from customer reviews by adding an LLM preprocessing step to the traditional topic-modeling pipeline. 
We improve the extraction of insights from customer reviews by restructuring the topic modelling pipeline to operate on opinion units -- distinct statements that include relevant text excerpts and associated sentiment scores. Prior work has demonstrated that such units can be reliably extracted using large language models. The result is a heightened performance of the subsequent topic modeling, leading to coherent and interpretable topics while also capturing the sentiment associated with each topic.
 By correlating the topics and sentiments with business metrics, such as star ratings, we can gain insights on how specific customer concerns impact business outcomes. %The use of opinion units addresses key challenges in prior work, particularly by isolating individual opinions within reviews creating coherent topic clusters. 
 %Our approach also enhances the interpretability of results, making it clear which sections of raw text influenced clustering or regression outcomes. 
We present our system's implementation, use cases, and advantages over other topic modeling and classification solutions. We also evaluate its effectiveness in creating coherent topics and assess methods for integrating topic and sentiment modalities for accurate star-rating prediction.
\end{abstract}

\section{Introduction}

Understanding customer feedback is important for businesses aiming to refine their operations, uncover growth opportunities, and align with customer needs. While many organizations collect substantial amounts of text-based input from customers, employees, and other stakeholders, the primary challenge lies not in the availability of information, but in extracting insights from large volumes of unstructured data \citep{tavakoli:2024}. %This article evaluates and exemplifies our proposed system using restaurant reviews. However, the solution is equally applicable to analyzing other types of opinionated text, such as product reviews, employee surveys, and patient feedback.

We propose the analysis framework TopicImpact, which integrates neural topic modeling with an LLM-enabled preprocessing step that extracts and structures opinions. 
%This article evaluates and exemplifies the system using restaurant reviews. 
In the preprocessing step, raw reviews are transformed by an LLM into opinion units \citep{HaglundBjorklund:2025a} -- extracted phrases that encapsulate a customer's sentiment on specific aspects. Additionally, the LLM assigns a sentiment score on a scale from 1 to 10, where~1 indicates a very negative sentiment and 10 represents a very positive one. (See Part A of Figure~\ref{system:desc} for examples of opinion units.)
By clustering these opinion units instead of entire reviews, TopicImpact generates more coherent and interpretable topic clusters. A detailed outline of the system is provided in Section \ref{system:desc}.
%Our paper introduces this novel application of opinion units, demonstrating their benefits in enhancing both topic modelling clarity and star rating prediction performance.

The improved topic clusters allow marketers to establish several important facts. By considering the themes of the generated clusters, they learn \emph{(1) the topics discussed in the reviews}, and by considering the cluster sizes, also \emph{(2) the prevalence of each topic}. Through the sentiment score of each opinion, they can infer \emph{(3) the sentiments associated with the topics}, and by applying regression to the the combination of topics and sentiments,  \emph{(4) the contribution of each topic to the business metric, both in terms of polarity (positive/negative) and degree}.
 By providing these four pieces of information, our solution contributes to a complete and detailed understanding of customer opinions in unstructured review text. 

%Avsnitt handlar om varför vår metod är bättre än traditionell klustring
The proposed framework offers clear advantages over traditional topic modeling methods. Unlike approaches that cluster entire reviews, TopicImpact generates more coherent topics by clustering aspect-delineated opinion units, this is an important strategy because individual reviews often address multiple aspects. Statistical methods like Latent Dirichlet Allocation (LDA) \citep{blei:2003} assume that documents are mixtures of words and, similar to TopicImpact, allow reviews to belong to multiple topics. However, LDA lacks the contextual understanding provided by embeddings created with pretrained language models. While LDA represents topics using keyword lists, it offers limited insight into which specific parts of a text prompted the assignment of a review to a topic. In contrast, TopicImpact enhances interpretability by using the opinion units' aspect-delineated label and excerpt as the clusterable document, offering clear context for understanding the topic assignment. Additionally, metadata links back to the original review. % This approach enhances interpretability and may build trust in the system.

%In contrast, opinion units extract phrases representing single opinions and has metadata that links back to the original review. This allows for a more granular understanding of topic assignment, as the excerpt containing the opinion serves as the document for clustering. This approach enhances interpretability and may build trust in the system.

%avsnitt handlar om vrf klustring över klassifikation
We can also relate topic modeling to classification, which is the appropriate approach when dealing with static and clearly defined categories. However, classification fails to capture emerging themes and niche customer concerns. Nor is classification suitable for  exploratory analysis — situations where we do not know what opinions to expect in advance. %In contrast, the self-organizing nature of topic modelling enables the detection of new topics, such as excitement over a new feature or dissatisfaction after a recent service change.  

Another drawback with classification is the difficulty of adapting the  level of topical abstraction. For example, shifting from analyzing general pricing concerns to examining a specific coupon code campaign would typically require retraining and relabeling datasets in traditional models. Even with LLMs that do not require extensive training data, this necessitates reclassification of all reviews, which incurs significant consumption of time, cost and compute. Topic modeling overcomes these limitations by enabling dynamic adjustments, such as setting the desired number of topics or providing seed words to refine the analysis toward specific topics. While our system also requires LLM-based preprocessing and embedding of opinion units—both of which are computationally demanding—these, are performed only once, enabling later rapid iteration and exploration of customer feedback across different levels of abstraction.

Beyond extracting insights from reviews, the applications of TopicImpact extend to other forms of opinionated free-text data, such as employee surveys, customer support interactions, course evaluations, and patient feedback. Instead of correlating topics with star ratings, we can analyse other business outcomes, such as employee or patient satisfaction, purchase likelihood or customer churn.

% \section{Why clustering over classification?}

% Clustering these opinion units provides a more comprehensive understanding of customer feedback by revealing the frequencies, intensities, and emerging trends of discussion topics. Unlike classification models, which require extensive labeled data and may overlook novel or unexpected themes, clustering offers a flexible and exploratory approach that adapts to the dynamic nature of customer opinions. This capability is crucial for identifying sudden changes in customer sentiment, such as the introduction of a highly popular feature or dissatisfaction with a recent service change.

%\section{Aim and Research Questions}

%This article presents the TopicImpact system, detailing its implementation and how LLM-based preprocessing enhances topic modeling to deliver new and improved capabilities for analyzing opinions in large volumes of text. We also describe the system's use cases and limitations. Additionally, we conduct experiments that exemplify the system's output and demonstrate its effectiveness by evaluating research questions on cluster quality and star rating prediction accuracy.   

In the upcoming sections, we outline the implementation of TopicImpact and discuss its intended use cases. We then conduct experiments to address the following key research questions.

\newcommand{\mkNewBox}[3]{
  \NewTotalTColorBox[auto counter]{#2}{ +m }{ 
    notitle,
    colback=#1!5!white,
    frame hidden,
    boxrule=0pt,
    enhanced,
    sharp corners,
    borderline west={4pt}{0pt}{#1},
  }{
    \textcolor{#1}{
        \sffamily
        \textbf{#3.}%
    }
    \ignorespaces
    ##1
  }
}

\mkNewBox{Rust}{\Definition}{Value Proposition}

\begin{description}
\item[RQ1:] Can topic modeling of opinion units generate coherent topics with regards to subject matter and sentiment? How does a sentiment-aware embedding model compare to a general-purpose embedding model in this context?

%\item[RQ1:] Can topic modeling of opinion units generate coherent topics that capture both topic and sentiment dimensions? Additionally, how does a sentiment-tuned embedding model compare to a general-purpose embedding model in this context?

\item[RQ2:] To what extent can topics generated through topic modeling accurately predict star ratings? Furthermore, how can the integration of sentiment and topic modalities enhance prediction accuracy?
\end{description}

%We demonstrate the effectiveness of our methodology through topic modeling techniques and semi-supervised clustering to group opinion units. This flexible approach allows businesses to adjust granularity based on their analytical needs, from broad categories to niche customer concerns. Our findings highlight the importance of clustering over classification in review analysis and showcase its potential to make a significant leap in an important NLP application.

\begin{figure*}[ht]
    \centering
    \includegraphics[width=\textwidth]{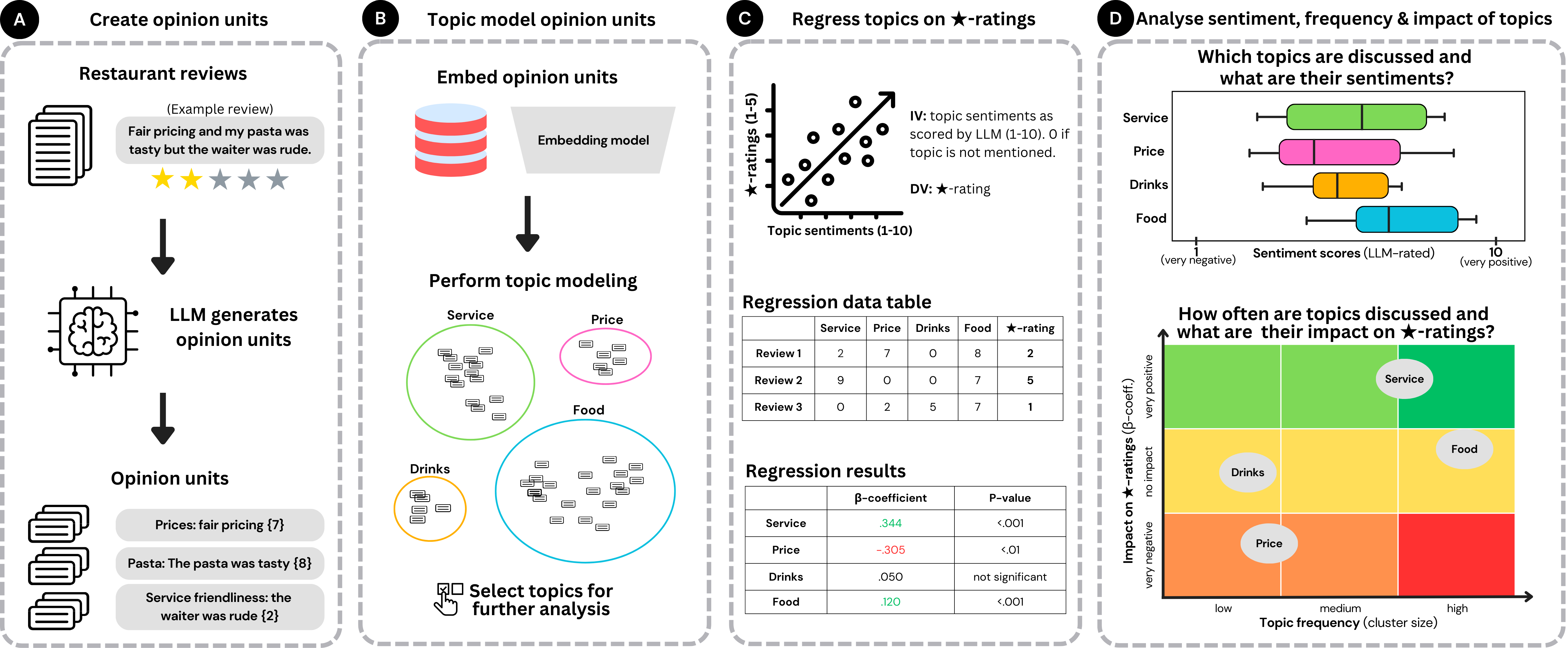}
    \caption{Overview of TopicImpact. \textbf{A:} LLM processes reviews to opinion units: \emph{label: excerpt \{sentiment 1-10\}}. \textbf{B:} Embed and cluster opinion units through topic modeling. \textbf{C:} Regress topics and sentiments onto star ratings. \\ \textbf{D:} Analyze sentiment, frequency, and topic impact.} 
    \label{fig:wide-image}
\end{figure*}

\Definition{
TopicImpact is a fast, interpretable and cost-effective solution for iteratively exploring customer opinions. It generates coherent topic clusters, provides supporting customer quotes, and identifies the sentiment associated with each topic, as well as its correlation to business metrics.} 

% \Definition{
% TopicImpact is a fast and cost-effective way to iteratively explore customer opinions. It generates coherent clusters, with extracted phrases that clarify the groupings,  and provide direct customer quotes. The ability to connect these phrases back to original reviews builds trust in the system.} 

%and understandable clusters, with the ability to trace extracted phrases back to their original reviews.}

\section{Related Work}
\label{sec:related}

%Aspect-bsed sentiment analysis
TopicImpact contributes to the field of aspect-based sentiment analysis \citep{zhang:2022}. The goal of this field is to understand which aspects are addressed in a text—such as a product review—and what sentiment the author expresses about each aspect. Early work treated the extraction of sentiments, terms, and categories in relative isolation \citep{liu:2015, li:2017, zhou:2015, luo:2019}. More recent studies have focused on extracting multiple factors simultaneously, capturing both the opinion aspect and its corresponding expression \citep{peng:2020, gao:2021}.

% Beskriv vår NodaLiDa artikel om word opinion units. Motivationen bakom opoinion units och resultatet som visar att LLMs kan skapa dem på ett "tillfredställande" sätt. 
Opinion units extend aspect-sentiment pairs by including excerpts from the target text that motivate the pairing. Compared to keywords, phrase extraction offer a more complete and nuanced representation of opinions. This added context benefits downstream tasks and, as we will show, supports clustering. Evaluations on review datasets demonstrate that LLMs can accurately extract opinion units using few-shot learning, with GPT-4 achieving a recall of 85.3\% and precision of 87.4\% when evaluated on restaurant reviews~\cite{HaglundBjorklund:2025a}. A key advantage of this approach is its flexibility: when it is not possible or desirable to adhere to a fixed aspect taxonomy, the LLM can define aspects on the fly. Opinion units have been evaluated for similarity search across academic and public review datasets, where it outperformed the traditional data segmentation strategies of sentence and passage chunking~\cite{HaglundBjorklund:2025a}.

In our work, we identify topics by embedding the opinion units and clustering the resulting vectors producing grouped themes. Embedding clustering works well for the short text excerpts involved, automatically determines the number of topics, and typically yields high topic coherence~\cite{grootendorst:2022}. The classical topic modeling alternative is Latent Dirichlet Allocation, which has a smaller computational footprint and produces interpretable topic representations through keyword distributions~\cite{Blei:2023}. A core principle of LDA is that documents are modeled as mixtures of topics, meaning each document is assumed to cover multiple topics. This is an important property for aspect-based sentiment analysis, where texts discuss opinions on several distinct aspects. Consequently, LDA and its variants have been adopted in unsupervised aspect-based sentiment analysis implementations~\citep{linshi:2014, debortoli:2016, krishnan:2023}.

A limitation of traditional embedding-based clustering, by contrast, is that each document is assigned to only one topic cluster. This restricts its ability to represent multi-aspect texts in which multiple opinions or themes coexist. At the same time, these methods benefit from the richer semantic representations of modern embeddings, producing clusters that—according to human evaluations—are more semantically coherent than those generated by LDA~\citep{churchill:2022}.  Our approach leverages the strengths of both methods by using LLM-preprocessing to segment texts into individual opinion units, each treated as a separate document for clustering. This mirrors LDA’s ability to assign multiple topics within a text, while benefiting from the richer semantic representations provided by embedding-based methods. Furthermore, LLM-preprocessing serves an additional function by filtering out non-opinionated content that is irrelevant for sentiment analysis~\cite{HaglundBjorklund:2025a}, removing noise from the topic modeling. 

LLMs can also be used for topic identification in a corpus without clustering ~\cite{Pham2024TopicGPTAP}, but \citet{li2025largelanguagemodelsstruggle} find that this often results in overly generic topics that do not aid understanding. 

% Topic Modeling: LDA vs document clustering with word embeddings

\section{System description}
\label{system:desc}
%The four main components of TopicImpact are: (A) extracting opinion units from raw reviews, (B) clustering these units into meaningful topics, (C) analyzing the relationship between the identified topics and business metrics such as star ratings, and (D) assessing the output to evaluate topic frequency, relevance, and importance in customer reviews. These steps are detailed in Figure 1 and the following section.  

The TopicImpact system is presented in Figure 1, through four steps (A-D).  
In step A, individual reviews are processed into opinion units using LLMs. These opinion units consist of an opinion label, a supporting excerpt, and a sentiment score (1–10), where 1 is very negative and 10 is very positive. An example review is processed in the figure. More details on the conceptualization of opinion units and an evaluation of LLM’s ability to generate them can be found in \citep{HaglundBjorklund:2025a}. %The prompt template used for generating opinion units for our article is found in the appendix. 
To isolate the contribution of specific topics, we instruct the LLM to tag overall sentiments towards the restaurant or experience (e.g., ``we had a wonderful time'') with an ``overall experience'' label. Since these statements are not informative for individual topics, we remove them from the analysis.

Step B starts by embedding the opinion units label and excerpt using a word embedding model, along with relevant metadata such as review ID and sentiment score. The opinion units are clustered through topic modeling, based on the widely used BertTopic~\cite{grootendorst2022bertopicneuraltopicmodeling}, with the number of clusters as a key parameter. The final step in this phase is to inspect the clusters and select topics for further analysis.

% The next step (B) involves embedding the opinion units using a word embedding model. Metadata such as the review ID, sentiment of the opinion, and the author ID are also included. The opinion units are then clustered using topic modeling (BertTopic). Key parameters at this stage include the desired number of clusters. %, with an optional seed for shaping the clusters. 
% The final step is to inspect the clusters and select topics of interest for further analysis. %Opinion units offer a significant advantage in inspecting formed clusters by extracting phrases tied to a specific opinion, along with metadata linking back to the original review. This enables a more granular understanding of why a review belongs to a particular cluster, as the full excerpt containing the opinion serves as the clusterable document—something that whole-document clustering or LDA cannot provide. This level of detail enhances interpretability and builds trust in the system.

In Step C, the relationship between the chosen topics and a business metric (here: star rating) is analyzed using multiple linear regression (MLR). Each review is a data point, with independent variables $X$ representing clustered topics, each in its own column. For example, if an opinion unit linked to a review is clustered under a `service' topic, the `service' column for this review data point is populated with the sentiment score of the opinion unit. If there are multiple mentions of `service' within the same review, an average sentiment score is calculated; if there are no mentions, the value is set to zero. The dependent variable $y$ is the star rating which ranges from 1 to 5.  This analysis provides coefficients for each topic, reflecting their strength of association with star ratings, along with $p$-values to assess statistical significance.  

% After selecting the topic, the next step is to analyze its relationship with a relevant business metric (in this case; star ratings) using multi-variate linear regression (see step C). Each review serves as a data point in the model, with the independent variables $X$ representing the clustered topics, each in its own column. If the review has an opinion unit clustered to a `service' cluster, then the `service' column is filled with the sentiment rating of the service-related opinion unit. If there are multiple mentions of `service' within the same review, an average sentiment score is calculated. If `service' is not mentioned, then value is set to zero. The dependent variable $y$ is the star rating (ranging from 1 to 5). This analysis provides the coefficients for each topic, indicating the strength of their relation to star ratings, as well as $p$-values, which allow us to determine whether this relationship is statistically significant. %In this article, we compare several methods for combining sentiment and topic modalities to predict star ratings as accurately as possible (see RQ1 and Method section).

In Step D we draw insights from the analysis. The topic clusters generated in Step B allow us to identify key topics discussed- their frequency, and sentiment. Regression analysis ranks topics by their relation to star-ratings. Combining this with frequency data highlights the topics most critical to customer satisfaction, enabling the creation of a priority matrix to rank issues by importance ~\citep{slack:1994}. For example, a frequently discussed topic that negatively affects satisfaction should be urgently addressed, while a positive but infrequent topic (e.g. sustainability) may present an opportunity for product promotion, such as highlighting locally-sourced ingredients.

%From topic modeling in step B we can answer the questions: What topics do our customer discuss, how often are they discussed and what are the sentiments within the topics? The distribution of sentiment can be visualized e.g. through a box-plot as in Section D - prices sentiments are overall negative while opinions about the food are positive. Our regression analysis enables us to rank the relative importance of various topics in relation to overall customer satisfaction. When combined with the frequency at which these topics are discussed, this analysis provides powerful insights into the key drivers of customer satisfaction. This results in a "priority matrix,"\citep{slack:1994} which helps identify which issues warrant immediate attention. For example, if a topic negatively impacts customer satisfaction and is frequently discussed, it should be prioritized for resolution. Conversely, if a topic has a positive impact but is infrequently mentioned, it presents an opportunity for further promotion. For example, emphasizing the use of locally-sourced ingredients more prominently in advertisements and within the restaurant could present an opportunity if this value proposition is currently under-promoted.

%\section{System description choices}

%\subsection{Regression Model}

{
\setlength{\parindent}{0cm}
\setlength{\parskip}{1ex}
\textbf{Regression Model.}\quad
Multiple linear regression is widely used in social science and marketing research for its interpretability and ability to assess the significance of relationships \citep{gordon:2015}. While non-linear models may improve prediction accuracy, they often compromise explainability. Despite criticisms of applying continous outcome regression to ordinal scales like star ratings or Likert scales \citep{binder:2019}, it remains the standard approach. Therefore, we adopt this method.
}
{
\setlength{\parindent}{0cm}
\setlength{\parskip}{1ex}
\textbf{Word Embeddings.}\quad
In our research questions, we ask whether coherent clusters of opinion units can be formed with respect to topic and sentiment. A key design consideration is the choice of word embedding. While general-purpose embeddings effectively capture semantic similarity, they often struggle to distinguish between opposing sentiments \citep{kim:2024}—e.g., ``the food is delicious'' and ``the food is disgusting'' might be considered similar due to shared topical context. Recent research addresses this by fine-tuning embeddings to incorporate sentiment information \citep{Tang2016SentimentEW, yu:2017-refining, fan:2022, kim:2024, ghafouri-etal-2024-love} We compare a general-purpose embedding, all-mpnet-base-v2 \citep{sent_transformers:2024}, with SentiCSE \citep{kim:2024}, a state-of-the-art, RoBERTa-based sentiment-aware model, to assess their impact on topic- and sentiment-cluster coherence. SentiCSE employs contrastive learning to separate positive and negative sentiment representations in embedding space. Through their LLM-rated sentiment metadata (1-10), opinion units offers an alternative approach to achieving sentiment-topic coherence by splitting data based on sentiment scores, then clustering using general-purpose embedding models. In our experiments, we compare these methods' ability to predict star ratings.
}
%\subsection{Unguided topic modelling vs guided topic modelling}

%Tror inte vi får plats med detta- vare sig att diskutera guided klustring ingående eller att bedriva experiment om det. 

%The main challenge in these tasks is the accurate pairing of aspect-sentiment elements \citep{zhang:2022}.

\section{Applications}
\label{sec:use_cases}
Our system offers distinct advantages for a range of use cases. We divide these into three classes:
\plparsep=.25ex
\plitemsep=.5ex
\setdefaultleftmargin{0ex}{1ex}{}{}{}{}
\begin{compactenum}
\item \emph{Exploratory Analysis of Customer Feedback.} Comparative analysis of product aspects or features: Understand which aspects resonate positively or negatively with customers. Evaluate how customer sentiments towards specific aspects of the business influence metrics like satisfaction, sales, churn, or brand loyalty. This customer knowledge could allow businesses to create more personalized promotions or experiences, tailored to specific customer segments' priorities and needs.

\item \emph{Identification of Emerging Trends and Issues. } 
Real-time tracking of emerging trends in customer feedback. Businesses can quickly detect responses to new features or pinpoint emerging issues.% following recent service changes. 

%It also is a big chance for targeted advertising, relying on impactful and new trends and creating opportunities for customer segmentation. 

\item \emph{Hypothesis-driven exploration}
For targeted investigations, the system enables a non-compute-intensive, iterative approach. For example, analysts can evaluate sentiment toward a new coupon campaign, refining parameters such as cluster size and using keyword-seeded clustering to explore this predefined topic of inquiry—without the compute and time costs of classification methods.

%For organizations with specific investigative goals, the system supports an iterative, non-compute-intensive approach. Teams can efficiently explore customer feedback, refine their exploration parameters, and through seeded clustering target specific questions without incurring the high costs associated with computationally expensive methods like classification. For example, this could involve analyzing sentiment towards a newly introduced coupon campaign. 
\end{compactenum}

TopicImpact is especially valuable for industries with large, diverse feedback data, such as e-commerce platforms and large retail chains, where customer preferences can vary widely and are often unpredictable. It is also beneficial for traditional service and product-based industries, including hotel and restaurant chains, fashion, automotive, or any business that gathers customer feedback. %The systems possible applications also extend to other forms of opinionated free-text data, such as employee surveys, customer support interactions, course evaluations, and patient feedback.

The demand for customer opinion analysis, is also evident in academic research. Several studies, particularly in the hotel, restaurant, and consumer-tech sectors, have focused on: (i) clustering opinions in reviews to analyze topic frequency and sentiment (e.g. \citep{linshi:2014, debortoli:2016, hu:2019, krishnan:2023}), and (ii) correlating topics with star ratings to evaluate their impact on overall business outcomes (e.g. \citep{debortoli:2016, linshi:2014, fu:2013, pappas:2014,ganu:2013, radojevic:2017, binder:2019}). These articles use methods such as LDA, keyword extraction, or available metadata, and therefore lack the advancements in explorability, cluster coherence and interpretability that LLM-preprocessing enables.

\section{Experiments}
We describe the experiments conducted to evaluate the proposed framework and research questions.

\subsection{Dataset and Preprocessing}

We base our experiments on the Yelp review dataset~\citep{yelp:2015}. Due to the its vast size, we restrict ourselves to reviews from US restaurants, and focus on three gastronomic styles—Italian, Mexican, and Japanese. The filtering of reviews is based on Yelp's metadata. This results in  three distinct datasets for topic modeling and star prediction analysis. Each dataset is subsampled to 5000 reviews.

Opinion units are generated using GPT-4, with a prompt  in Appendix A1. We exclude opinion units related to the overall experience, focusing instead on specific aspects as described in Section \ref{system:desc}. On average, each review generates 5.65 opinion units.

\subsection{Topic Modeling}
\label{sec:clustering_setup}

For our experiments, we adopt the BERTopic pipeline outlined by \citep{grootendorst:2022}, employing hard HDBSCAN \citep{campello:2013} for clustering and  UMAP for dimensionality reduction. We consider two alternative sentence embedding models: a general-purpose model \texttt{all-mpnet-base-v2} \citep{sent_transformers:2024} and a state-of-the-art sentiment-aware embedding model (\texttt{SentiCSE}; see \citep{kim:2024}). For our evaluation, we set the number of topics to 20 to ensure a manageable workload for human evaluation and the minimum topic size to 50 to provide sufficient data for statistical significance in regression analysis. 

We then evaluate the coherence of the topic clusters and their predictive ability for star ratings. To accurately reflect system performance, we include all topics in the analysis without any sub-selection. However, subselection is often helpful in applications to focus on the topics that are of greatest interest to the user (discussed in Sections \ref{system:desc} and \ref{sec:use_cases}).  

\subsection{Evaluation of Topic Modeling}

We evaluate the coherence of topics clusters with regards to topic and sentiment. For each cluster, three evaluators identify a dominant topical theme and determine which opinion units align with this theme (an inclusion-based approach \citep{eklund:2024}). The precision of a cluster is defined as the proportion of opinion units that belong to its dominant theme. Evaluators were assigned 20 randomly sampled opinion units per topic. Further details on the evaluation are provided in Appendix A2. 

To assess evaluator consistency, 5 opinion units per topic were reviewed by all evaluators, allowing us to calculate inter-rater agreement as ($m/n$), where $m$ is the number of overlapping opinion units on which all evaluators agreed, and $n$ is the number of total overlapping units.

Sentiment precision is calculated based on the distribution of LLM-assigned sentiment scores. We define it as the percentage of opinions with the dominant sentiment (positive or negative). For example, if 18/20 opinion units in a cluster have a sentiment score $>5$ (positive), the precision is 90\%. The same applies if 18/20 opinion units are negative (sentiment score $\le 5$ ).

%Furthermore, we complement our analysis with a traditional keyword-based measure of cluster coherence, \emph{$c_v$}, which quantifies the statistical co-occurrence of words within a cluster \citep{roder:2015}.

\subsection{Star Prediction: Regression analysis} 
\label{sec:star_prediction}
After the topic modelling has been applied to the extracted opinion units, we can predict the star-rating of an individual review $r$ through multiple linear regression (MLR). The basis for the regression is the opinion units extracted from $r$— their sentiment scores and their topic membership assigned through the topic modeling described in Section \ref{sec:clustering_setup}. The regression model is given in Equation~\ref{eq:star_rating}, where $K$ is the number of topics and $s_{k}(r)$ represents the average sentiment scores of the opinion units belonging to $r$ that occur in topic cluster $k$ (see Figure 1 and Section 2). %to for each topic in Review $k$.
\begin{equation}
\scalebox{1.5}{$\star$}\text{-rating}(r) = \beta_0 + \sum_{k=1}^{K} \beta_k \cdot s_k(r) 
\label{eq:star_rating}
\end{equation}

%Each review is treated as a data point, with sentiment scores for opinion units within a topic being averaged if multiple units from the same review are clustered together (see Figure 1 and Section 2).

%In the regression, review is treated as a single data point, with the sentiment scores of the opinion units in each topic/cluster serving as the s-parameters. If multiple opinion units from the same review are clustered together, their sentiment scores are averaged.

% \begin{align}
% \scalebox{1.5}{$\star$}\text{-rating} &= \beta_0 + \beta_{\text{topic1}} s_{\text{topic1}} + \beta_{\text{topic2}} s_{\text{topic2}} \nonumber \\
% &\quad + \dots + \beta_{\text{topick}} s_{\text{topick}} + \varepsilon \label{eq:star_rating}
% \end{align}

We implement three different methods for integrating topic and sentiment information to predict star ratings and compare their performance. To control for the effect of clustering granularity, we vary the number of clusters $K \in \{10, 20, 30\}$. For each method, we conduct evaluations both with sentiment scores (as defined in Equation~\ref{eq:star_rating}) and without sentiment scores. It the latter case, we let $s_k(r)=1$ if topic $k$ is mentioned in review $r$ and 0 otherwise.% This yields a total of six experiment groups.

\plparsep=.25ex
\plitemsep=.25ex
\setdefaultleftmargin{0ex}{.5ex}{}{}{}{}
\begin{description}
    \item[M1.] Cluster with a general embedding model, here \texttt{all-mpnet-base-v2} \citep{sent_transformers:2024}.
    \item[M2.] Cluster using a sentiment-aware embedding model, here SentiCSE \citep{kim:2024}.
    \item[M3.] Split opinion units into negative and positive data based on LLM ratings (negative: $\leq$ 5, positive: $> 5$), then perform topic modeling separately using the general embedding model. %(\texttt{all-mpnet-base-v2}). 
\end{description}
For Method 3, the sentiment scores (1-10) are adjusted to a 1-5 scale. For the cluster based on the positive data split, ratings of 1-5 correspond to the previous 6-10 range on the 1-10 scale, while for the negative cluster, the ratings of 1-5 remain. %If a review does not mention a particular topic, that topic is as previously assigned a value of 0.
Splitting the data before clustering means that positive and negative influences are  weighted separately. For instance, good service might not significantly impact the star rating because it is often expected, whereas poor service can be a decisive factor in lowering the overall experience.  

We evaluate the predictive performance of the regression models using $R^2$ and RMSE on a hold-out sample with 5-fold cross-validation.

\section{Results}
The alternative combinations of clustering approach and embedding model are evaluated based on (i) cluster coherence and (ii) effectiveness in predicting star ratings.

\subsection{Topic and Sentiment Coherence}

\begin{comment}
\begin{table}[t]
\centering
\setlength{\tabcolsep}{3.0 pt}  % Reduce column spacing
\scriptsize
%\begin{tabular}{|l|l|c|c|c|c|c|}
\begin{tabular}{llccccc}
\toprule
\multirow{2}{*}{\textbf{Embedding}} & \multirow{2}{*}{\textbf{Dataset}} & \multirow{2}{*}{\textbf{Outliers \%}} & \multicolumn{2}{c}{\textbf{Topic (P)}} & \multicolumn{2}{c}{\textbf{Sentiment (P)}} \\
 & & & Mean & $\ge$.9 & Mean & $\ge$.9 \\
\midrule
\multirow{3}{*}{\begin{tabular}[c]{@{}l@{}}General-\\Purpose\end{tabular}} & Italian & 24.9\% & 90.5\% & 79.0\% & 76.1\% & 20.0\% \\
 & Mexican & 16.9\% & 91.7\% & 79.0\% & 73.4\% & 5.0\% \\
 & Japanese & 29.3\% & 86.3\% & 63.2\% & 73.0\% & 15.0\% \\
\addlinespace[0.3em] % adds a little space before the hline
\hline
\addlinespace[0.3em]
\multirow{3}{*}{\begin{tabular}[c]{@{}l@{}}Sentiment-\\Aware\end{tabular}} & Italian & 22.9\% & 84.3\% & 55.3\% & 90.8\% & 75.0\% \\
 & Mexican & 32.0\% & 83.2\% & 60.5\% & 86.6\% & 60.0\% \\
 & Japanese & 20.1\% & 82.1\% & 52.6\% & 90.2\% & 70.0\% \\
\bottomrule
\end{tabular}
\caption{Topic and sentiment precision for cluster coherence, showing average precision and the percentage of clusters with precision $\ge$ 90\%, K=20 clusters.}
\label{tab:cluster_precision}
\end{table}
\end{comment}

\begin{table}[t]
\centering
\setlength{\tabcolsep}{3.0 pt}  % Reduce column spacing
\scriptsize
%\begin{tabular}{|l|l|c|c|c|c|c|}
\begin{tabular}{llccccc}
\toprule
\multirow{2}{*}{\textbf{Embedding}} & \multirow{2}{*}{\textbf{Dataset}} & \multirow{2}{*}{\textbf{\% Outliers  }} & \multicolumn{2}{c}{\textbf{Topic (P)}} & \multicolumn{2}{c}{\textbf{Sentiment (P)}} \\
 & & & Avg. & $\ge$.9 & Avg. & $\ge$.9 \\
\midrule
\multirow{3}{*}{\begin{tabular}[c]{@{}l@{}}General-\\Purpose\end{tabular}} & Italian & 24.9  & .905  & 79.0  & .761  & 20.0  \\
 & Mexican & 16.9  & .917  & 79.0  & .734  & 5.0  \\
 & Japanese & 29.3  & .863  & 63.2  & .730  & 15.0  \\
\addlinespace[0.3em] % adds a little space before the hline
\hline
\addlinespace[0.3em]
\multirow{3}{*}{\begin{tabular}[c]{@{}l@{}}Sentiment-\\Aware\end{tabular}} & Italian & 22.9  & .843  & 55.3  & .908  & 75.0  \\
 & Mexican & 32.0  & .832  & 60.5  & .866  & 60.0  \\
 & Japanese & 20.1  & .821  & 52.6  & .902  & 70.0  \\
\bottomrule
\end{tabular}
\caption{Average topic and sentiment precision for cluster coherence, \% of clusters with precision $\ge .9$ and \% of outliers (opinion units not assigned to a cluster by BERTopic) for each embedding-dataset pair ($K=20$).}

%\caption{Percentage of outliers for each combination of embedding and dataset, followed by topic and sentiment precision for cluster coherence (showing average precision) and the percentage of clusters with precision $\ge .9$ for $K=20$ clusters. \% Outliers is the percentage of opinion units that BERTopic did not assign to a cluster.}
\label{tab:cluster_precision}
\end{table}

%For the general embedding model(\texttt{all-mpnet-base-v2}) the average topic precision is between X\% of topics have a $90\%$ precision demonstrating an ability to create a high topic coherence \citep{eklund:2022}, see Table \ref{tab:cluster_precision}. The sentiment-aware model (\texttt{sentiCSE}) performs consistenly worse across all three datasets. The results also show that the general-purpose embedding model does not pay much attention at all to the sentiment polarity of opinions when clustering, the sentiment aware precision does however creating multiple clusters on the same topic theme but opposing sentiment e.g. a positive price cluster and a negative one. This focus on sentiment is also one explanation for why the topic precision worse is: as it produces topics that are sentiment but not topically coherent- e.g. topics that say that aspects are "ok" whether it be service, food quality or cleanliness.    

For the general-purpose embedding model (that is, \texttt{all-mpnet-base-v2}), the average topic precision over the clusters for each dataset fall in the range 86.3-91.7\%, with 63.2-79.0\% of topics achieving 90\% precision (see Table \ref{tab:cluster_precision}), demonstrating a high topic coherence \citep{eklund:2022}. The sentiment-aware model (\texttt{sentiCSE}) consistently performs worse across all three datasets. The results suggest that the general-purpose model largely ignores sentiment polarity when clustering, whereas the sentiment-aware model creates multiple clusters for the same topic with opposing sentiment (e.g., positive vs. negative price clusters). This focus on sentiment partly explains the lower topic precision, as it forms clusters that are cohesive in terms of sentiment but inconsistent in terms of topic. For instance, opinions about service, food quality, and cleanliness may be grouped together simply because they were all described as ``okay''.

Inter-rater agreement among evaluators was 90.3\%. For topics with a clear theme (e.g., precision $\ge$80\%), agreement was 96.1\%. 
%
%Inter-rater agreement among evaluators was 90.9\%, and even higher when excluding incoherent topics with no clear theme. For topics with a clear theme (e.g., those with precision ≥80\%) inter-rater agreement was 96.3%. 
%
%Our results show that topic modeling with a general embedding model (\texttt{all-mpnet-base-v2}) consistently across the three datasets outperforms a sentiment-aware model (\texttt{sentiCSE}) with regards to creating cohesive topic clusters, see Table \ref{tab:cluster_precision}. The results also show that the general-purpose embedding model does not pay much attention at all to the sentiment polarity of opinions when clustering, the sentiment aware precision does however creating multiple clusters on the same topic theme but opposing sentiment e.g. a positive price cluster and a negative one. This focus on sentiment is also one explanation for why the topic precision worse is: as it produces topics that are sentiment but not topically coherent- e.g. topics that say that aspects are "ok" whether it be service, food quality or cleanliness. 
%
The percentage of outliers not assigned to a cluster ranges from 17-32\%. While BERTopic can optionally force documents into clusters, we believe that having outliers is valid, as some opinions naturally fall outside the larger groups. Therefore, we allowed outliers. 

To answer RQ1, our topic modeling results—comparing one general and one sentiment-aware embedding— achieve high topic cohesion but not high sentiment precision simultaneously. However, the metadata of opinion units can be leveraged to create sentiment coherence. We implement this in the next section, where one of our star-rating prediction method splits the data based on sentiment-scores before topic modeling.

%To answer RQ1, topic modeling can achieve high topic cohesion but not simultaneously as high sentiment precision. However, the metadata of opinion units can be leveraged to create sentiment coherence. We implement this in the next section, where one of our methods for star-rating prediction involves splitting the dataset based on sentiment-ratings before clustering.

%To answer RQ1, topic modeling can achieve high topic cohesion but not simultaneously as high sentiment precision. However, the metadata of opinion units can be leveraged to create sentiment coherence. We implement this in the next section, where one of our methods for star-rating prediction involves splitting the dataset based on sentiment-ratings before clustering.

\subsection{Star Prediction: Regression Analysis}
\label{sec:results_regression}

In Table \ref{tab:star_prediction}, we present the results of star-rating predictions using the three methods outlined in Section~\ref{sec:star_prediction}. The regression results show that sentiment-aware embeddings (Method 2) yield more accurate star rating predictions than general embeddings (Method 1). This improvement is expected, as topic clusters are formed based on both sentiment and topic, providing more relevant information about the customer’s overall satisfaction. Method~3, which splits the dataset based on LLM-sentiment scores into positive and negative opinion units before clustering each split separately, achieves the highest accuracy with an $R^2$ value of 0.726, indicating a strong model fit. When clustering with general embeddings, incorporating sentiment scores of the individual opinions as described in Section \ref{sec:star_prediction}, rather than using one-hot encoding for topic membership, enhances prediction accuracy. 

The results are consistent across datasets and numbers of clusters. Table \ref{tab:star_prediction} shows regression outcomes averaged over the three datasets, with $5$-fold cross-validation (K=20 clusters). Detailed results for each dataset and number of clusters ($K=10, 20, 30$) are provided in Appendix A3.

To answer RQ2, our results show that TopicImpact accurately predicts star ratings. Topic modeling alone using general embeddings yields unsatisfactory results due to insufficient sentiment capture. However, combining topic assignment with the opinion units' sentiment scores—by segmenting data by sentiment before clustering and incorporating sentiment scores into regression (Eq.~\ref{eq:star_rating})—achieves high regression accuracy.

% \begin{table}[h!]
% \centering
% \scriptsize
% \begin{tabular}{|l|l|c|c|c|} 
% \hline
% \textbf{Method} & \textbf{$s_i$-ratings} & \textbf{Sig. Coeffs. (\boldmath{$\beta$})} & \textbf{R2} & \textbf{RMSE} \\ \hline
% Method 1 & Without & 14.5 &.113 & 1.31 \\ 
% & With & 14.6 &.383 & 1.12 \\ 
         
% Method 2 & Without & 14.5& .487 & 1.01 \\ 
% & With & 15.3 & .393 & 1.10 \\ 
         
% Method 3 & Without & 11.5 & .652 & .825 \\ 
%          & With & 15.1 & \textbf{.726} & \textbf{.731} \\ 
% \hline
% \end{tabular}
% \caption{Star-prediction: \( R^2 \) and RMSE on holdout sample (\( K = 20 \) clusters), values averaged over datasets.}
% \label{tab:star_prediction}
% \end{table}

\begin{table}[h!]
\centering
\scriptsize
\begin{tabular}{@{}lcccc@{}} 
\toprule
\addlinespace[0.2em]
\textbf{Method} & \textbf{$s_k$ scores} & \textbf{\# Sig. \boldmath{$\beta$}} & \textbf{$R^2$} & \textbf{RMSE} \\ 
\addlinespace[0.1em] \midrule \addlinespace[0.2em]
\multirow{2}{*}{\begin{tabular}[c]{@{}l@{}}M1. General Embeddings\end{tabular}} & without & 14.5 &.113 & 1.31 \\ 
& with & 14.6 &.383 & 1.12 \\ 
\addlinespace[0.1em] \hline \addlinespace[0.2em]       
\multirow{2}{*}{\begin{tabular}[c]{@{}l@{}}M2. Sentiment-Aware\\ Embeddings\end{tabular}} & without & 14.5& .487 & 1.01 \\ 
& with & 15.3 & .393 & 1.10 \\ 
\addlinespace[0.1em] \hline \addlinespace[0.2em]   
\multirow{2}{*}{\begin{tabular}[c]{@{}l@{}}M3. Sentiment Splitting\end{tabular}} & without & 11.5 & .652 & .825 \\ 
         & with & 15.1 & \textbf{.726} & \textbf{.731} \\ 
\bottomrule
\end{tabular}
\caption{Star-prediction: No. of significant $\beta$-coefficients (out of 20),  \( R^2 \) and RMSE (\( K = 20 \)).}% , values averaged over datasets.}
\label{tab:star_prediction}
\end{table}

%\subsection{Example output from TopicImpact} 

To illustrate the output from TopicImpact, we present results from the Japanese restaurant dataset using Method 3: Sentiment Splitting, with $K = 20$ clusters (10 negative and 10 positive). In Table \ref{tab:top_clusters_full} we summarize the topics ranked by their influence on star rating. For each topic, the table includes the beta coefficient, the topic size (number of assigned opinion units), and a representative opinion unit.

% Below in Table \ref{tab:top_clusters_full} we present results from the japanese dataset using Method 3: Sentiment Splitting, with K = 20 clusters (10 negative and 10 positive). Below we summarize the topics ranked by their influence on star ratings (coefficients). 

% \begin{table}[ht]
% \centering
% \scriptsize
% \setlength{\tabcolsep}{2.0 pt}
% \begin{tabular}{@{}lccl@{}}
% \toprule
% \textbf{Topic} & \textbf{\boldmath{$\beta$}} & \textbf{Size} & \textbf{Opinion Unit Example} \\
% \midrule
% \cellcolor[HTML]{FFD9D9} & \cellcolor[HTML]{FFD9D9} & \cellcolor[HTML]{FFD9D9}53,2\% & \cellcolor[HTML]{FFD9D9} Food safety \\
% \cellcolor[HTML]{FFD9D9}\multirow{-2}{*}{49\%} & \cellcolor[HTML]{FFD9D9}\multirow{-2}{*}{49\%} & \cellcolor[HTML]{FFD9D9} Food poisoning: The next day was spent in the & \cellcolor[HTML]{FFD9D9}52,2\% \\
% \bottomrule
% \end{tabular}
% \end{table}

\begin{table}[ht]
\centering
\scriptsize
\setlength{\tabcolsep}{2.0 pt}
\begin{tabular}{@{}lrrp{.64\linewidth}@{}}
\toprule
\textbf{Topic} & \multicolumn{1}{c}{\textbf{\boldmath{$\beta$}}} & \textbf{Size} & \textbf{Opinion Unit Example} \\ 
%\midrule
%\multicolumn{4}{c}{\emph{Top 5 Positive Topics}} \\
\midrule
 \rowcolor{red!10}% 
\rowcolor{red!10}  Food safety & -1.21 & 56 & Food poisoning: The next day was spent in the bathroom for several hours getting rid of the sushi \\[.3ex]
\rowcolor{red!10} Service & -.886 & 1718 & Staff friendliness: Our waiter had a very unfriendly standoffish personality\\
 \rowcolor{red!10} Food & -.852 & 2398 & Food quality: My food was awful... overcooked \\
 \rowcolor{red!10} Order error & -.621 & 76 & Order accuracy: I got the complete wrong order\\
  \rowcolor{red!10} Portions  & -.511 & 73 & Portion size: The serving was pathethically small \\
\rowcolor{red!10} Hibachi & -.458 & 363 & Hibachi grill entertainment: the chef was lacking in entertainment. I was borderline bored \\
\rowcolor{red!10} Pricing & -.346 & 755 & Sushi pricing: Pricing was too high for a conveyor belt style. These shops are popular in Japan, Taiwan, and China yet they were charging \$11.00 a roll compared to other country's \$1.50 a roll\\
\rowcolor{red!10} Cleanliness & -.311 & 921 & Cleanliness: the restaurant was filthy\\
\rowcolor{red!10}  Drinks & -.215 & 170 & Sake selection: it tasted like water with a little vodka poured into it\\
\rowcolor{red!10} Parking & -.161  & 121 & Parking: A negative is the parking spots are narrow. It would deter me from going there when it's busy \\
  
%\midrule
%\multicolumn{4}{c}{Top 5 Negative Topics} \\
\midrule
\rowcolor{green!10} Food & .835 & 7182 & Sushi: The sushi is always spot on delicious \\
\rowcolor{green!10} Service & .350 & 1623 & Service: the service is super good. The waitresses will check on you and do refills \\
\rowcolor{green!10} Price & .347 & 1268 & Food value: the food is very good for the price \\
\rowcolor{green!10} Atmosphere & .346 & 182 & Atmosphere: Casual and comfortable atmosphere! \\
\rowcolor{green!10} Interior & .229 & 758 & Décor: The decor is a simple and clean aesthetic \\
\rowcolor{green!10} Noodles & .137 & 72 & Nabeyaki udon: The udon chicken tastes AMAZING. it hands down the best nabeyaki udon in town\\
\rowcolor{green!10} Busyness & .137 & 248 & Restaurant crowding: Whenever we've gone on a weekend evening, it's never been terribly crowded\\
\rowcolor{green!10} Family friendly & n.s. & 75 & Suitability for families: great for families! \\
\rowcolor{green!10} Location & n.s. & 337 & Location: The location of this place is actually beautiful \\
\rowcolor{green!10} Views & n.s. & 66 & Views: The place has wonderful views, since it's right on the river\\
\bottomrule
\end{tabular}
\caption{\emph{Topic} is a description of the topic, \emph{$\beta$.} is the regression coefficient in star rating prediction, \emph{Size} is the number of opinion units in the topic, and the final column provides an example opinion unit. "n.s." indicates that the $\beta$-coefficient was not statistically significant (i.e., p > .05).}
\label{tab:top_clusters_full}
\end{table}

\section{Conclusion}

TopicImpact enhances the extraction of actionable insights from customer reviews by integrating topic modeling with LLM-powered segmentation of reviews into distinct opinion units—individual, separated opinions supported by text excerpts. This provides both topic coherence and interpretability, while the metadata of opinion units enables sentiment-based segmentation. By incorporating sentiment scores and correlating topics with business metrics, TopicImpact effectively predicts star ratings and reveals how customer concerns influence business outcomes. This approach offers a fast, cost-effective solution for businesses to explore and act on customer feedback.

%Include both limitations with TopicImpact system and limitations with the evaluation of the system.

A promising avenue for future research is integrating keyword or example-guided topic seeding within our framework of topic modeling with LLM preprocessing. This approach could combine the predefined segmentation of classification with the exploratory, iterative, and cost-effective benefits of topic modeling. Furthermore, verifying the performance of TopicImpact in other domains, such as retail product reviews, course evaluations, or employee satisfaction surveys, could improve opinion analysis in these critical areas. The different domains could pose distinct challenges, including varied opinion lengths, increased opinion nuance, or more frequent non-opinion content. Strategies to tackle these issues might involve customizing prompts, designing fine-tuning datasets, and thoughtfully combining abstractive and extractive summarization techniques to produce clear opinion units.

\section{Limitations}

%For accurate topic segmentation, where the desired topics are known beforehand, classification methods should be used instead of clustering, as performance will be much higher.  Clustering is a valuable option when the priority is exploration, iterativeness, and cost-effectiveness. A promising alternative is keyword or example-guided clustering, which combines the predefined segmentation of classification with the advantages clustering. Demonstrating its potential in this domain offers a compelling area for future work.

A limitation of LLM preprocessing is that it sometimes misses opinions in reviews or creates excerpts lacking full context \citep{HaglundBjorklund:2025a}. While the fidelity of extracted opinion units is high using GPT-4 \citep{HaglundBjorklund:2025a} some errors persist, which can undermine the accuracy of subsequent opinion analysis. However, preprocessing also addresses key issues in opinion clustering, such as removing non-opinionated text.

In this work, we evaluate our system's ability to generate coherent topics and predict star ratings by comparing a general-purpose embedding model (all-mpnet-base-v2) with a sentiment-aware embedding (sentiCSE). While the comparison reveals clear trends between the general and sentiment-aware embeddings, further validation using a larger number of embedding models would enhance the reliability and generalizability of these conclusions.

Another limitation of this work is the lack of a standardized benchmark dataset for evaluating aspect-based sentiment analysis in the context of topic modeling. Benchmark datasets such as 20 Newsgroups are commonly used to assess topic models in domains like news \citep{churchill:2022}. However, traditional aspect-based sentiment analysis datasets, such as the SemEval restaurant reviews \citep{pontiki2016semeval}, consist of short, sentence-length entries that fail to capture the complexity of full-length reviews. As a result, they are ill-suited for evaluating topic models applied to real-world review data, which is the focus of our work. To address this gap, we use authentic, full-length reviews and rely on human evaluation. This enables a context-sensitive assessment of both topic relevance and sentiment—better reflecting real-world use cases. Although evaluating topic models using a labeled dataset has its issues—particularly the difficulty of aligning fixed abstraction levels with an unsupervised task—such evaluation could complement our results and support benchmarking against alternative approaches.

%The system is not fully automated in that manual inspection of clusters is still advisable, though not strictly necessary, to ensure they are homogeneous and at the desired level of abstraction. However, this process often contributes to understanding customer feedback, making it a valuable step in many cases.

Another set of limitations arises not from our system evaluation, but from the inherent challenges of topic modeling itself, particularly when chosen over classification methods. When the desired topics are known in advance, classification is typically the more appropriate approach, as it offers higher performance for accurate topic segmentation. However, topic modeling is more suitable for exploratory tasks where flexibility, iteration, and cost-effectiveness are priorities. One key issue with topic modeling is the potential misalignment between generated topics and user expectations. For example, should a topic such as ``tiramisu'' form its own cluster, or would it be better grouped under a broader ``dessert'' category or even inside a general ``food'' cluster? The behavior and granularity of topic formation can be influenced by settings such as the number of clusters or the use of seeded clustering techniques. However, the inability to exert complete control over topic granularity and assignment is a drawback for applications that require stable and pre-defined topic categories.

%Another set of limitations, do not relate to our system evaluation but rather the use topic modeling in general, particularly to the choice of such methods over classification. For accurate topic segmentation, where the desired topics are known beforehand, classification methods should be used instead of clustering, as performance will be much higher.  Clustering is a valuable option when the priority is exploration, iterativeness, and cost-effectiveness. An example of the the issues with topic modelings is that topics may not always align with expectations. For instance, should a "tiramisu" cluster exist independently, or would it be more appropriately placed within a broader dessert category—or even a general food cluster? The clustering behavior and level of abstraction can be influenced by parameters such as cluster size or seeded clustering. However, the effectiveness of these approaches, which can be highly subjective and this subjectivity remains an inherent issue with topic modeling \citep{}. 

% \section{Limitations}


\bibliography{references}

\newpage
\appendix

\section{Appendix}

\subsection{Prompt template}

\begin{figure}[h]
\noindent

\input{prompt_template}
  \caption{Prompt template for generating opinion units.}
  \label{prompt_template}
\end{figure}

%\newpage

\subsection{Instructions for Evaluators Analyzing the Clustered Topics}
\label{sec:appendix}

The evaluation of topic clusters was performed by one of the authors and 2 unpaid volunteers recruited through personal outreach. They were provided the following instructions.

\begin{compactenum}
    \item Select the Topic: Click on a topic in the topic list.
    \item Review the Excerpts: Carefully read through the 20 opinions/restaurant review excerpts belonging to the selected topic.
    \item Identify a Common Theme: Look for a common theme among the review excerpts. Focus only on the topic being discussed (e.g., service, desserts, cleanliness, etc.). Ignore the sentiment of the opinions—whether they are positive or negative is not relevant.
    \item Name the Topic: In the Excel file, write a descriptive name for the topic in the ``Topic Name'' column. If there is no clear majority theme, choose the most common or representative theme based on your judgment.
    \item Flag Errors: Identify any review excerpts that DO NOT FIT the topic. In the Excel file, list the IDs of these excerpts under the ``Error IDs'' column. Separate the IDs with commas (e.g., 5,7,15).
\end{compactenum}

\subsection{Additional Star Rating Regression Results.}

The following tables present the detailed star rating regression results as discussed in Section \ref{sec:results_regression}. These results are shown for each dataset (Italian, Mexican, and Japanese) and when varying number of clusters (K) used in the topic modeling. Results are averages over 5-fold cross validation on a hold-out sample. 

\begin{table}[h!]
\centering
\scriptsize
\setlength{\tabcolsep}{5.0 pt}
%\begin{tabular}{|l|l|c|c|c|c|}
\begin{tabular}{llcccc} 
\hline
\addlinespace[0.2em]
\textbf{Method} & \textbf{$s_k$-ratings} & K & \textbf{R2} & \textbf{RMSE} & \textbf{Sig. Coeffs. (\boldmath{$\beta$})} \\ 
\addlinespace[0.1em] \hline \addlinespace[0.2em]
% Add your data rows here
\multirow{6}{*}{\begin{tabular}[c]{@{}l@{}}M1. General\\ Embeddings\end{tabular}} & With & 10 & .482 & 1.02 & 6.2/10 \\ 
 & Without & 10 & .090 & 1.35 & 8.6/10 \\ 
  & With & 20 & .380 & 1.12 & 12.6/20 \\ 
   & Without & 20 & .136 & 1.32 & 16.0/20 \\ 
  & With & 30 & .335 & 1.16 & 20.2/30 \\ 
   & Without & 30 & .145 & 1.31 & 21.2/30 \\ 
\addlinespace[0.1em] \hline \addlinespace[0.2em] 
\multirow{6}{*}{\begin{tabular}[c]{@{}l@{}}M2.Sentiment-\\Aware\\ Embeddings\end{tabular}} & With & 10 & .366 & 1.13 & 8.8/10 \\ 
 & Without & 10 & .441 & 1.07 & 9.4/10 \\ 
  & With & 20 & .378 & 1.13 & 17.2/20 \\ 
   & Without & 20 & .463 & 1.05 & 17.0/20 \\ 
  & With & 30 & .350 & 1.16 & 25.2/30 \\ 
   & Without & 30 & .456 & 1.06 & 25.4/30 \\ 
\addlinespace[0.1em] \hline \addlinespace[0.2em] 
\multirow{8}{*}{\begin{tabular}[c]{@{}l@{}}M3. Sentiment\\Splitting\end{tabular}} & With & 10 & .766 & .687 & 8.0/10 \\ 
 & Without & 10 & .671 & .814 & 8.6/10 \\ 
 & With & 20 & .750 & 712 & 14.4/20 \\ 
 & Without & 20 & .678 & .808 & 15.4/20 \\ 
  & With & 30 & .728 & .744 & 20.0/30 \\ 
   & Without & 30 & .661 & .832 & 21.0/30 \\ 
  & With & 40 & .716 & .763 & 31.0/40 \\ 
   & Without & 40 & .666 & .827 & 31.0/40 \\ 
 \hline
% Add more rows as needed
\end{tabular}
\caption{Italian restaurant dataset: Star prediction results (\( R^2 \) and RMSE) and no. of significant $\beta$-coefficients, varying the number of clusters (\( K \)).}
\end{table}

\begin{table}[h!]
\centering
\scriptsize
\setlength{\tabcolsep}{5.0 pt}
\begin{tabular}{llcccc} 
\hline
\addlinespace[0.2em]
\textbf{Method} & \textbf{$s_k$-ratings} & K & \textbf{R2} & \textbf{RMSE} & \textbf{Sig. Coeffs. (\boldmath{$\beta$})} \\ 
\addlinespace[0.1em] \hline \addlinespace[0.2em]
\multirow{6}{*}{\begin{tabular}[c]{@{}l@{}}M1. General\\ Embeddings\end{tabular}} & With & 10 & .544 & .900 & 7.2/10 \\ 
 & Without & 10 & .089 & 1.27 & 8.2/10 \\ 
  & With & 20 & .364 & 1.06 & 14.8/20 \\ 
   & Without & 20 & .119 & 1.25 & 14.8/20 \\ 
  & With & 30 & .335 & 1.09 & 19.0/30 \\ 
   & Without & 30 & .136 & 1.25 & 17.8/30 \\ 
\addlinespace[0.1em] \hline \addlinespace[0.2em] 
\multirow{6}{*}{\begin{tabular}[c]{@{}l@{}}M2.Sentiment-\\Aware\\ Embeddings\end{tabular}} & With & 10 & .385 & 1.05 & 8.6/10 \\ 
 & Without & 10 & .503 & .945 & 7.2/10 \\ 
  & With & 20 & .396 & 1.05 & 12.2/20 \\ 
   & Without & 20 & .531 & .922 & 9.8/20 \\ 
  & With & 30 & .401 & 1.04 & 21.2/30 \\ 
   & Without & 30 & .531 & .918 & 20.0/30 \\ 
\addlinespace[0.1em] \hline \addlinespace[0.2em] 
\multirow{8}{*}{\begin{tabular}[c]{@{}l@{}}M3. Sentiment\\Splitting\end{tabular}} & With & 10 & .720 & .705 & 8.8/10 \\ 
 & Without & 10 & .641 & .797 & 8.8/10 \\ 
 & With & 20 & .705 & .726 & 15.6/20 \\ 
 & Without & 20 & .635 & .807 & 14.8/20 \\ 
  & With & 30 & .697 & .736 & 23.4/30 \\ 
   & Without & 30 & .642 & .802 & 21.8/30 \\ 
  & With & 40 & .695 & .743 & 29.0/40 \\ 
   & Without & 40 & .641 & .806 & 28.4/40 \\ 
 \hline
% Add more rows as needed
\end{tabular}
\caption{Mexican restaurant dataset: Star prediction results (\( R^2 \) and RMSE) Star prediction results (\( R^2 \) and RMSE) and no. of significant $\beta$-coefficients, varying the number of clusters (\( K \)).}
\end{table}

\begin{table}[H]
\centering
\scriptsize
\setlength{\tabcolsep}{5.0 pt}
\begin{tabular}{llcccc} 
\hline
\addlinespace[0.2em]
\textbf{Method} & \textbf{$s_k$-ratings} & K & \textbf{R2} & \textbf{RMSE} & \textbf{Sig. Coeffs. (\boldmath{$\beta$})} \\ 
\addlinespace[0.1em] \hline \addlinespace[0.2em]
% Add your data rows here
\multirow{6}{*}{\begin{tabular}[c]{@{}l@{}}M1. General\\ Embeddings\end{tabular}} & With & 10 & .456 & 1.06 & 8.8/10 \\ 
 & Without & 10 & .049 & 1.40 & 8.0/10 \\ 
  & With & 20 & .326 & 1.18 & 16.4/20 \\ 
   & Without & 20 & .084 & 1.37 & 12.8/20 \\ 
  & With & 30 & .320 & 1.18 & 20.2/30 \\ 
   & Without & 30 & .125 & 1.34 & 20.6/30 \\ 
\addlinespace[0.1em] \hline \addlinespace[0.2em]
\multirow{6}{*}{\begin{tabular}[c]{@{}l@{}}M2.Sentiment-\\Aware\\ Embeddings\end{tabular}} & With & 10 & .498 & 1.02 & 6.6/10 \\ 
 & Without & 10 & .533 & .985 & 7.6/10 \\ 
  & With & 20 & .404 & 1.12 & 16.6/20 \\ 
   & Without & 20 & .466 & 1.05 & 16.8/20 \\ 
  & With & 30 & .422 & 1.1 & 27.6/30 \\ 
   & Without & 30 & .482 & 1.04 & 25.4/30 \\ 
\addlinespace[0.1em] \hline \addlinespace[0.2em]
\multirow{8}{*}{\begin{tabular}[c]{@{}l@{}}M3. Sentiment\\Splitting\end{tabular}} & With & 10 & .727 & .750 & 9.6/10 \\ 
 & Without & 10 & .640 & .861 & 10.0/10 \\ 
 & With & 20 & .724 & .754 & 15.2/20 \\ 
 & Without & 20 & .643 & .859 & 17.2/20 \\ 
  & With & 30 & .691 & .800 & 22.8/30 \\ 
   & Without & 30 & .627 & .880 & 22.2/30 \\ 
   & With & 40 & .688 & .807 & 29.6/40 \\
& Without & 40 & .629 & .880 & 30.2/40 \\ 
 \hline
% Add more rows as needed
\end{tabular}
\caption{Japanese restaurant dataset: Star prediction results (\( R^2 \) and RMSE) and no. of significant $\beta$-coefficients, varying the number of clusters (\( K \)).}
\end{table}

\end{document}